\documentclass[sn-mathphys-num]{sn-jnl}

\usepackage{caption}
\usepackage[utf8]{inputenc}
\usepackage{subcaption}
\usepackage{graphicx}%
\usepackage{multirow}%
\usepackage{amsmath,amssymb,amsfonts}%
\usepackage{amsthm}%
\usepackage{mathrsfs}%
\usepackage[title]{appendix}%
\usepackage{xcolor}%
\usepackage{textcomp}%
\usepackage{manyfoot}%
\usepackage{booktabs}%
\usepackage{algorithm}%
\usepackage{algorithmicx}%
\usepackage{algpseudocode}%
\usepackage{listings}%

\theoremstyle{thmstyleone}%
\newcommand\major[1]{{{#1}}}
\theoremstyle{thmstyletwo}%

\theoremstyle{thmstylethree}%

\begin{document}

\title[Article Title]{Enhancing Human-Robot Collaboration: A Sim2Real Domain Adaptation Algorithm for Point Cloud Segmentation in Industrial Environments}


\author*[1,3]{\fnm{Fatemeh} \sur{Mohammadi Amin}}\email{mohm@zhaw.ch}

\author[2]{\fnm{Darwin} \sur{G. Caldwell}}

\author[3]{\fnm{Hans Wernher} \sur{van de Venn}}

\affil[1]{\orgname{University of Zurich}, \city{Zurich}, \country{Switzerland}}

\affil[2]{ \orgname{Italian Institute of Technology (IIT)}, \orgaddress{\city{Genoa}, \country{Italy}}}

\affil[3]{\orgdiv{Institute of Mechatronics Systems}, \orgname{Zurich University of Applied Sciences}, \orgaddress{\city{Winterthur}, \country{Switzerland}}}


\abstract{The robust interpretation of 3D environments is crucial for human-robot collaboration (HRC) applications, where safety and operational efficiency are paramount. Semantic segmentation plays a key role in this context by enabling a precise and detailed understanding of the environment. Considering the intense data hunger for real-world industrial annotated data essential for effective semantic segmentation, this paper introduces a pioneering approach in the Sim2Real domain adaptation for semantic segmentation of 3D point cloud data, specifically tailored for HRC. 
Our focus is on developing a network that robustly transitions from simulated environments to real-world applications, thereby enhancing its practical utility and impact on a safe HRC.

In this work, we propose a dual-stream network architecture (FUSION) combining Dynamic Graph Convolutional Neural Networks (DGCNN) and Convolutional Neural Networks (CNN) augmented with residual layers as a Sim2Real domain adaptation algorithm for an industrial environment. 
The proposed model was evaluated on real-world HRC setups and simulation industrial point clouds, it showed increased state-of-the-art performance, achieving a segmentation accuracy of 97.76\%, and superior robustness compared to existing methods.}



\maketitle

\section{Introduction}\label{sec:introduction}

Safe human-robot collaboration (HRC) is becoming increasingly important under the Industry 4.0 paradigm, which emphasizes intelligent and flexible automation. 
Traditional robots are being replaced by collaborative robots (cobots), designed for safe and efficient interaction with humans in dynamic industrial settings.
In fact, cobots are increasingly being used for flexible task accomplishment instead of traditional industrial robots~\cite{olender2019cobots} and can work in the same workspace as humans~\cite{vicentini2021collaborative}.

\major{
To effectively collaborate with humans, cobots need advanced intelligence, allowing them to respond to human inputs and adapt to dynamic environments for smooth, safe, and productive workflows~\cite{mohammadi2020mixed}. This requires advanced sensing and perception capabilities~\cite{hamon2020robustness}, like semantic segmentation of 3D point cloud data, which is vital for accurately understanding objects in industrial environments and ensuring safety by detecting potential hazards~\cite{rangnet2019, guo2018review}.}

Despite the advances in machine learning techniques, deploying these models in industrial settings poses significant challenges. This is particularly true for tasks that involve humans, such as human-robot collaboration, where ensuring human safety is crucial~\cite{mohammadi2020mixed}. Factors such as data hunger—a term referring to the performance limitations due to insufficient datasets-, restrictions on using realistic datasets, and the time-consuming nature of manual annotation prevent the acquisition of large-scale, labeled point cloud datasets, which are necessary for training accurate models\cite{Hungry}.
\major{To address these challenges, transfer learning and domain adaptation techniques, which enable models to generalize across diverse datasets and domains, are increasingly being explored~\cite{surveyTrans}.}

This paper addresses the urgent issue of semantic segmentation of 3D point clouds within HRC environments. 
It achieved this by leveraging Sim2Real domain adaptation techniques, where the use of real-world data is not completely feasible due to safety concerns and practical limitations such as equipment availability constraints like the lack of diverse robots, and data collection difficulties. 
Sim2Real domain adaptation refers to strategies aimed at adapting a model trained in a simulated environment to perform efficiently in the real world. Therefore, the utilization of simulation data as an alternative means to train robust and efficient semantic segmentation models was explored. This leveraging of simulation data not only overcomes the lack of annotated real-world datasets but also provides a controlled environment to generate diverse and challenging scenarios for comprehensive model training.

One of the primary challenges associated with employing simulation data for training semantic segmentation models lies in achieving data realism and accuracy.
By leveraging cutting-edge simulation tools, it is possible to generate synthetic 3D point clouds that closely mimic industrial environments, while maintaining control over crucial aspects such as object interactions, lighting conditions, and material properties.

\major{A two-stream network called FUSION is leveraged in the methodology for effective segmentation. This approach combines two distinct neural network architectures to enhance performance, optimizing the system for more accurate segmentation tasks.
This network is a crucial component of the proposed approach. Initially, training is conducted using simulated data. Subsequently, the network is refined and adapted through real-world data applications, ensuring its robust performance and accurate interpretation within the dynamic environments of real-world industrial settings.}

The results showcase enhanced generalizability and performance in Human-Robot Collaboration environments, achieving 97.6 percent accuracy in real-world settings and 98.9 percent in simulation scenarios. 
The results also underscore the potential of this approach to contribute substantially to human safety in industrial Human-Robot Collaboration (HRC) settings by ensuring accurate and reliable identification of human and related safety hazards.
The main contributions of this paper are as follows:
\begin{itemize}
    \item A domain adaptation algorithm is presented to segment a 3D sparse point cloud.
    \item The dataset for domain adaptation, containing both, real and simulation data, is published.
    \item This is expected to be the first domain adaptation algorithm for real-time and real HRC applications, with testing conducted in real-world laboratory setups.
\end{itemize}

The rest of this paper is laid out as follows. Section II provides a review of related work in the areas of semantic segmentation of 3D point cloud data and Sim2Real domain adaptation. Section III describes the proposed methodology, while Section IV presents the experimental results. Finally, Section V offers a discussion of the findings and implications, along with potential directions for future research.

\begin{figure*}[t] 
\centering
\includegraphics[width=1\textwidth]{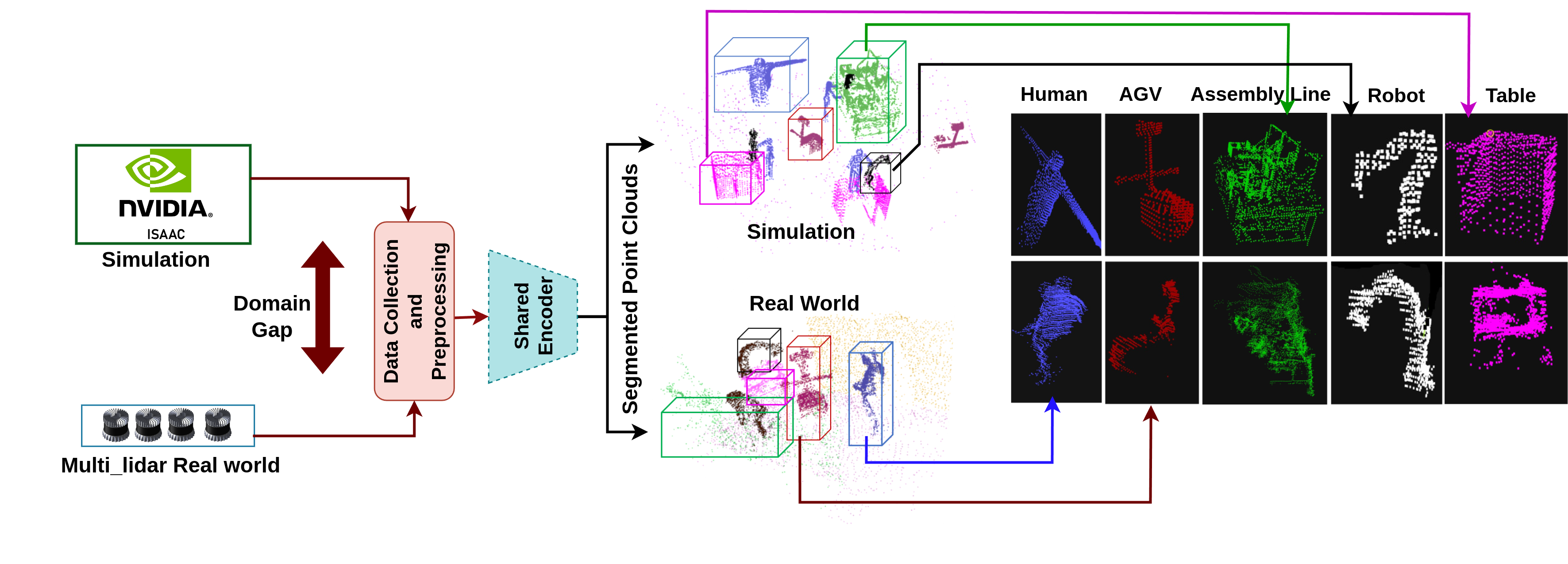}
\caption{The Overal concept of FUSION, Simulation and Real-world Dataset}
\label{fig:simReal}
\end{figure*}

\section{Related Works}\label{sec:relatedworks}
\subsection{\textbf{Semantic Segmentation of point clouds: Techniques and Challenges for HRC}}
Semantic segmentation of 3D LiDAR (Light Detection and Ranging) point clouds, which are often sparse and disordered~\cite{casc2020}, aims to assign a semantic label to every point. The most common strategies~\cite{guo2020deep} for this task are projection based~\cite{Multi1, MultiRGBD, wu2019squeezesegv2, milioto2019rangenet}, voxel-based~\cite{Voxel2017,VV-NetVoxel}, and point-based~\cite{Shellnet, Rec3dseg, Landrieu_2018_CVPR} methods. Deep learning has shown significant progress in research on this topic, particularly with PointNet~\cite{pointnet2016} and PointNet++~\cite{pointnetplus} architectures that directly process raw point cloud data. Further advancements like PointCNN~\cite{pointcnn} and DGCNN~\cite{DGCNN} introduced convolution-like operations, enhancing local structures and geometrical capture. 
However, training of these models mainly relies on ideal simulated data, hence their real-world applicability can be limited due to the domain shift. 

Point cloud representation, which maintains the original geometric information in 3D space without discretization, has shown substantial interest in real-world applications, including autonomous driving~\cite{Salsa2019,squeez2017,Rt3d2018,salsanext2020,S3net2021,thomas2019kpconv,rangnet2019}, navigation~\cite{auto2018} and robotics~\cite{robot2020} applications. Directly converting point clouds into sparse vector representations is a point-based approach that combines neighboring features for each point. Several studies within this domain have achieved impressive performance on various benchmark datasets. KPConv operators~\cite{thomas2019kpconv} and the RandLA-Net network~\cite{hu2020randla} are notable for their good performance in large-scale point cloud segmentation. They also improve memory efficiency and computational resources through reduced data redundancy and adaptive sampling techniques.
\major{Despite these advancements, the focus on semantic understanding in HRC applications has been insufficient, with most HRC research~\cite{imagescene, ANGLERAUD2024102663} concentrating on image-based data.
Recent developments in HRC focus on both perception and control mechanisms, as well as the integration of Human Factors (HF) to improve safety and efficiency in collaborative environments.} 

\major{In the perception part, previous works~\cite{Multimodal,multimodal3,multimodal2} provide comprehensive reviews of multimodal human-robot interaction, emphasizing the integration of various sensory inputs such as vision, auditory, and tactile feedback to enhance situational awareness and interaction quality. Also, Lin et al.~\cite{10471885} considered the integration of prior information alongside simulation data to improve point cloud segmentation, offering valuable insights into enhancing semantic segmentation accuracy. While this multimodal data fusion holds promise for complex scene recognition, improving real-time processing, robustness, and safety are critical challenges that still need to be addressed.}

\major{In this context, recent studies focus on integrating Human Factors (HF) into segmentation models to enhance the safety and robustness of these systems~\cite{HumanFactors, DENOBILE2024103663}. HF ensures that segmentation models focus on critical areas relevant to safety~\cite{HumaninLoop, HFinSafety}, such as identifying humans or hazardous objects in shared workspaces~\cite{HF2}.}
On the other hand, the lack of precise 3D information is a significant shortcoming as it is crucial for accurately detecting object location, which in turn is vital to ensure safety in human-robot interactions~\cite{mohammadi2020mixed}. This deficiency can lead to potential injuries~\cite{Jiang1987ACA}, necessitating rigid industrial robotic cell designs to ensure safety standards~\cite{9196924}. The lack of industrial benchmark datasets is a key drawback in this research area.

Moreover, collecting and annotating large datasets for each new task is costly, time-consuming, and impractical for scalability. Typical scenarios often involve detection tasks in one domain using training data from another domain, resulting in potential disparities in feature space or data distribution. The research community has proposed several techniques to facilitate knowledge transfer between domains to counteract performance degradation, with most methods focusing on 2D camera image domain adaptation techniques~\cite{surveyTrans}. 

In the proposed HRC segmentation task in this paper, Unsupervised learning was not used due to the complexity and specificity of the task, which may require labeled data to ensure the model learns relevant features and patterns directly related to the segmentation of human and robot interactions~\cite{HRCSupervised}.

\subsection{\textbf{LiDAR based Datasets}}

The advancements in LiDAR technology and its increasing implementation in domains like robotics, autonomous vehicles, and environmental research have been tremendous.
As part of these developments, datasets like the S3DIS~\cite{s3dis}, Semantic3D~\cite{semantic3d}, and SemanticKITTI~\cite{behley2019semantickitti}, designed for autonomous vehicles, and large-scale indoor and outdoor modeling respectively, emerged. 
However, a noticeable gap in the datasets specifically designed for Human-Robot Collaboration (HRC) environments remained evident. Until recently, a publicly available dataset tailored for such environments did not exist.

In a significant move to bridge this gap, We introduced the COVERED dataset  \cite{covered}.
The COVERED dataset, using multiple LiDAR sensors, made significant strides in addressing the needs of HRC environments. It captures a variety of interaction scenarios between humans and robots. These scenarios encompass interaction, observation, passing by, and autonomous operation. The dataset incorporates elements such as the dynamics of the collaboration, variability in human behavior, robot and AGV configurations, and more.

The COVERED dataset has limitations, mainly due to human safety and ethical restrictions that prevent the execution of certain hazardous scenarios in a real environment, which are critical for effectively training robotic systems. To address this, the use of simulated datasets has been considered. Simulation provides a safe platform to study potentially dangerous situations without risk, offering a controlled environment for systematic variable analysis.

Simulated environments combined with real-world data from multi-LiDAR sensors constitute a powerful solution for preparing robots for HRC tasks. The COVERED dataset offers a broad spectrum of realistic scenarios, while simulation allows for the investigation of edge cases, flexible scenarios, and potential hazards. Together, they make more reliable and safer robotic systems for human-robot collaboration.
 
 \subsection{\textbf{Domain Adaptation of Point Clouds}}
The concept of Sim2Real domain adaptation, a specific form of transfer learning, has emerged as a solution to the aforementioned issue and to mitigate performance degradation when applying a model to a different domain, a phenomenon known as domain shift. Domain adaptation aims to bridge this gap between the source and target domain. This can be categorized into three major types based on the availability of data from the target and source domains \cite{surveyTrans}: 
\begin{itemize}
 \item unsupervised transfer learning, where no annotated data is used; 
 \item transductive transfer learning, where annotated data is only available in the source domain; 
 \item inductive transfer learning, where annotations are available in the target domain.
 \end{itemize}

Despite advances in Sim2Real computer vision techniques enhancing model performance from synthetic to real images, Sim2Real domain adaptation for 3D point cloud semantic segmentation in industrial settings is still widely unexplored

Various researchers have proposed unique approaches to address this domain gap. Li et al. \cite{CompDom} developed a domain adaptation approach to bridge the gap in 3D point clouds acquired with different LiDAR sensors for autonomous driving. Jia et al. \cite{DBLP:MobGap} studied the domain gap between autonomous driving and mobile robot applications to enhance mobile robot navigation in human-populated environments. Langer et al. \cite{langer} proposed a real-to-real domain adaptation technique (different sensor data) for semantic understanding in situations lacking labeled data for a new sensor setup.

Additionally, Zhao et al. \cite{epointDa} introduced ePointDA, an end-to-end simulation-to-real domain adaptation (SRDA) framework, for LiDAR point cloud segmentation to bridge the domain shift between simulation image-based data and real LiDAR data. Yi et al. approached the LiDAR sampling domain gap for cars and pedestrians by treating the issue as a surface completion task.
Lastly, Jiang and Saripall \cite{LiDARNet} proposed a boundary-aware domain adaptation approach for autonomous driving tasks, introducing a new LiDAR dataset, SemanticUSL, which includes diverse scenarios such as traffic-road, walkpath, and off-road scenes, captured by a mobile robot.

In industrial settings, semantic segmentation of 3D point cloud data has crucial applications in process optimization and robotics \cite{respointnet2, chen2024pointdcunsupervised, wu2023sim2real}.

Despite all this mentioned research, the application of these techniques to enhance 3D point cloud data semantic segmentation in industrial contexts and HRC is still emerging.
This paper seeks to build upon this body of work, further exploring the effectiveness of transfer learning and domain adaptation in addressing the data hunger problem specific to industrial semantic segmentation of 3D point cloud data.
These techniques have been thoroughly explored in this paper, focusing on simulation scenarios and real-world evaluations, aiming to develop robust, accurate, and efficient models for industrial applications.

\section{Methodology}\label{sec:Dataset}

In this work, a Sim2Real domain adaptation approach was introduced to enhance semantic understanding within Human-Robot Collaboration (HRC) environments. \major{Figure \ref{fig:simReal} illustrates the example dataset of both the real world and simulation. It is evident that the quality of the objects in the simulation is significantly higher than in the real world. }This methodology leverages a dual-stream network architecture (Called FUSION) to analyze point clouds in parallel, as depicted in Figure \ref{fig:architecture}. Initially, a simulated dataset closely mirroring real-world industrial settings was created. This simulation serves as the primary training ground for the network, enabling it to learn and adapt to the real environment. Domain adaptation techniques are then applied to transition the model's learning from the simulated dataset to real-world applications. The network's performance is rigorously evaluated for both online and offline segmentation.
\begin{figure}[]
    \centering
    \begin{subfigure}[b]{1\textwidth}
        \centering
        \includegraphics[width=\textwidth]{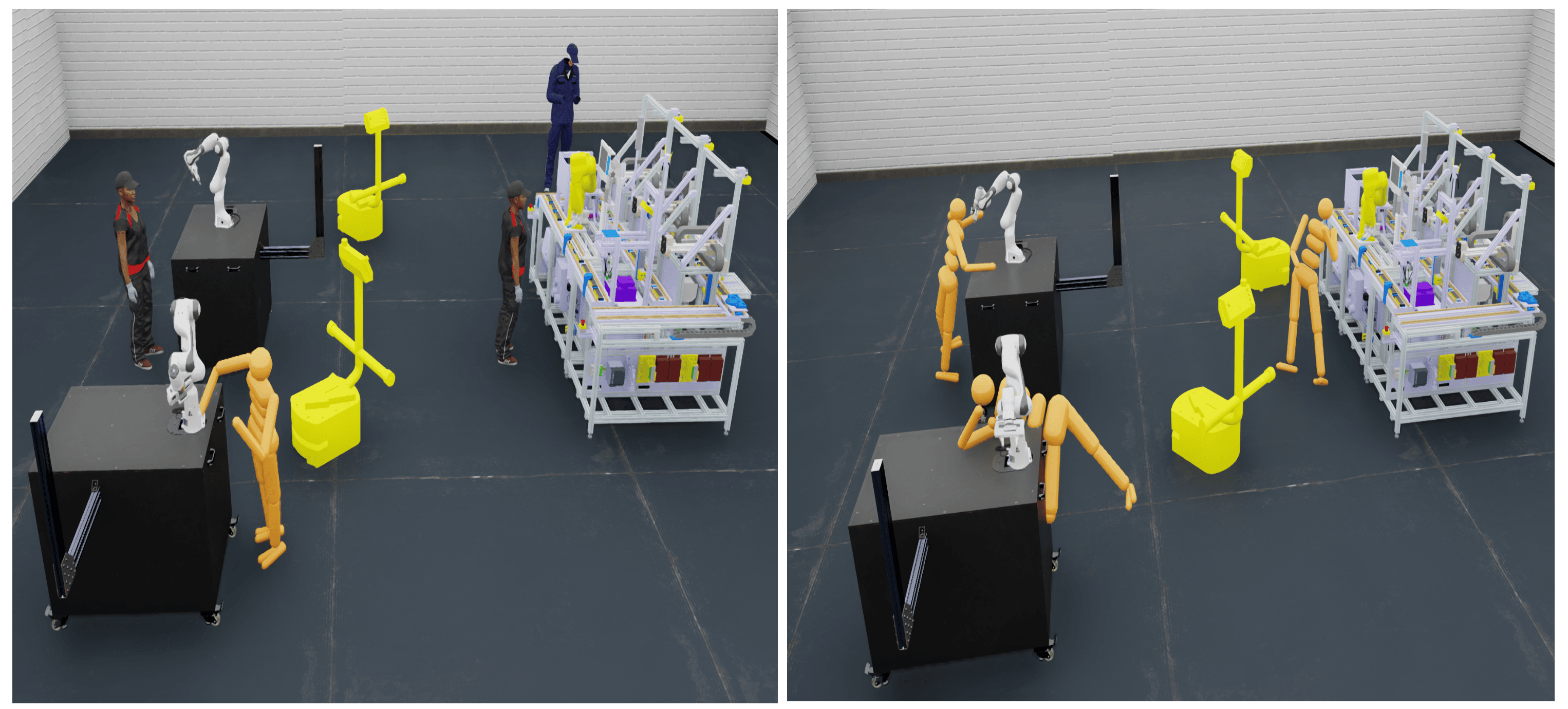}
        \caption{Safe Interaction vs Dangerous Interaction}
        \label{fig:first_image}
    \end{subfigure}
    \hfill 
    \begin{subfigure}[b]{1\textwidth}
        \centering
        \includegraphics[width=\textwidth]{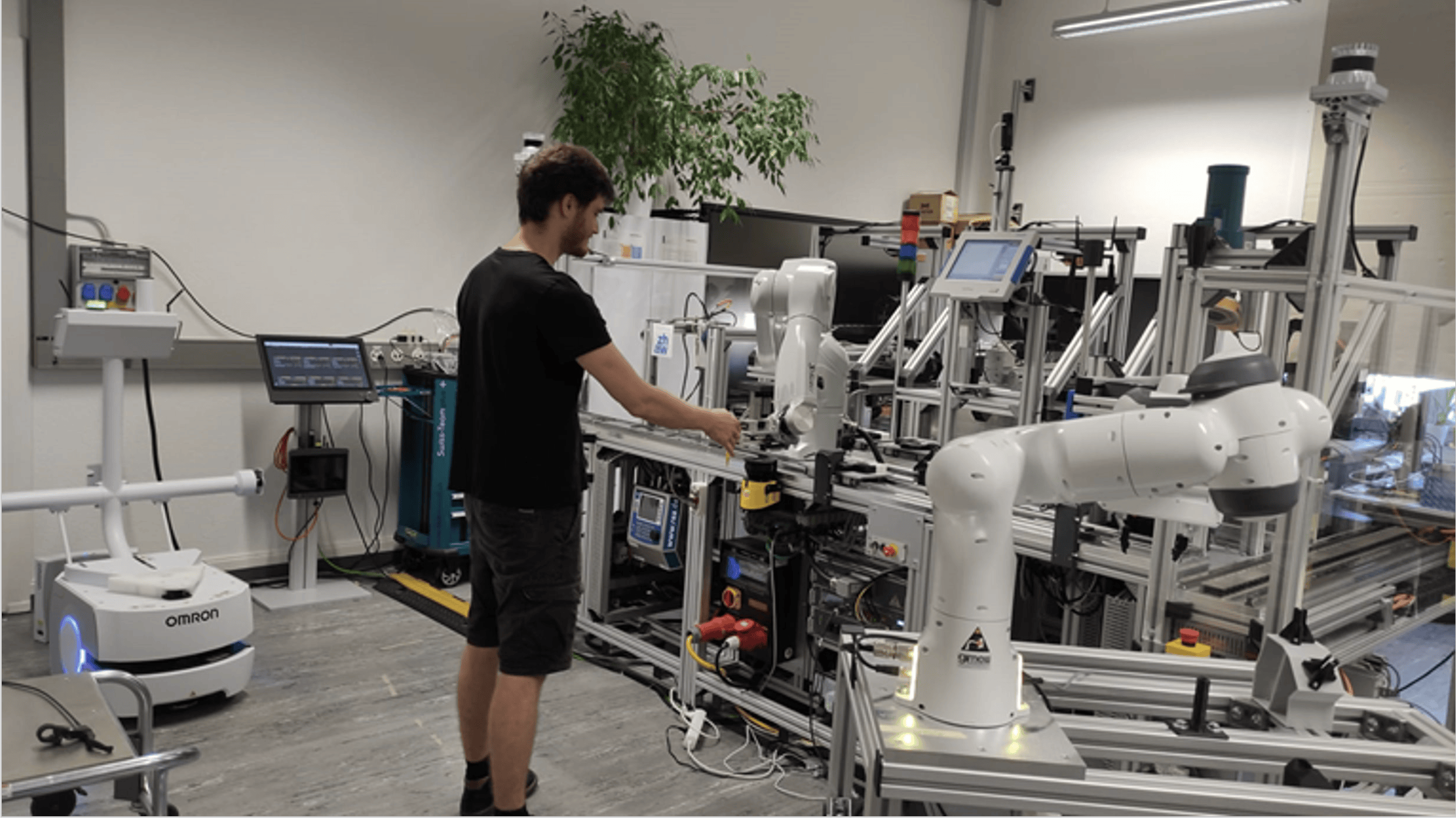}
        \caption{Real World}
        \label{fig:real}
    \end{subfigure}
    \caption{Simulation and Real-world setup}
    \label{fig:dataset_images}
\end{figure}
\subsection{\textbf{Synthetic Dataset}}
Simulation platforms have made it possible to meet inherent challenges in robotics, particularly the acquisition of diverse, real-world, annotated data.
In this study, a synthetic point cloud dataset was created using IsaacSim, a sophisticated robotics simulation platform by NVIDIA\cite{IsaacSim}. 
Diverse and realistic scenarios were generated by varying multiple parameters in the simulation, such as object poses and configurations. The robot models, AGVs, and human operator representations were placed in a variety of positions and orientations, and their interactions were carefully chosen to show possible real-world scenarios in industrial environments.
Potentially dangerous scenarios, which may be infrequent or absent in real-world datasets, were also created. This dataset aims to simulate certain hazardous situations that are restricted by safety regulations in real-world environments.

The synthetic dataset is comprised of 300 point clouds, each reflecting a different interaction scenario among robots, humans, and environmental elements. These represent a variety of real-world scenarios that may arise in a human-robot collaboration environment.

Notably, this synthetic dataset has been carefully structured to align with our previously published real-world dataset COVERED \cite{covered}.
The initial release of the COVERED dataset included 215 data samples categorized into 5 distinct classes: 
\textit{'Floor', 'Wall', 'Robot', 'Human'} and\textit{'AGV'}.
In the latest dataset annotation update, we have introduced two additional classes, \textit{'Assembly Line'} and \textit{'Table'}, along with approximately 350 more data samples, further enriching the dataset's diversity and applicability. 
These enhancements are reflected in the updated version of the COVERED dataset available on our GitHub repository \footnote{\href{https://github.com/Fatemeh-MA/COVERED}{https://github.com/Fatemeh-MA/COVERED}}. 

For a comprehensive understanding of the real-world data collection methods and setup, it is recommended to review the paper cited as \cite{covered}, which outlines the procedures and technologies used to gather the data forming the basis of the target domain.
The collected synthetic dataset, designed to replicate these real-world conditions, serves as the source domain for this study. By aligning the synthetic data with the real-world conditions, the dataset's utility for domain adaptation strategies was enhanced.

The creation of this synthetic dataset highlights the dedication to strengthening the availability of high-quality resources in the field of robotics research. The goal is to make this dataset a valuable tool for the development, testing, and improvement of algorithms in various areas of robotics, such as perception, navigation, and human-robot interaction. In addition, by aligning it with our real-world dataset, the synthetic dataset presents a distinct opportunity to advance research in domain adaptation and address the industrial data hunger challenges, potentially enhancing both model performance and human safety (by detecting dangerous interactions). This advancement will greatly support the transition of robotic applications from controlled simulation environments to real-world scenarios. Figure \ref{fig:dataset_images} illustrates the similarity between the synthetic dataset generated from simulation and the real-world setup and it shows the potential for seamless integration of simulated data into real-world applications.

\subsection{\textbf{Data Preprocessing}}

The proposed preprocessing stage involved several crucial steps applied uniformly to both source (synthetic) and target (real-world) point cloud datasets.
First, the datasets were zero-centered and normalized, bringing all input features into the same numerical range. This process is critical for neural network training as it accelerates convergence and prevents any single feature from disproportionately influencing the model due to its scale.
Next, hyperparameter tuning was conducted to optimize network performance. Using a systematic search over a predetermined range of values, the best-performing hyperparameters on a validation set was identified. 

Efforts were made to determine the best k value for the K-nearest neighbor (KNN) algorithm and explored how point cloud complexity and density relate to this optimal k value. 
Through testing, it was found that the optimal k value was influenced by the complexity and density of the point cloud. 

\major{Specifically, in the COVERED dataset, which consisted of point clouds with an average density of 11000 points and varying levels of complexity, a lower \(k\) value like \(k=10\) consistently outperformed higher values like \(k = 20\)  in terms of segmentation accuracy and it has shown to be more effective by focusing on the most relevant neighboring points. }

Subsequent to the generation of the initial dataset, data augmentation techniques were employed to increase its richness and diversity. Techniques such as random flipping, scaling, rotation, and noise injection were used to create variations in the data. Each augmented sample retained the labels from its original version, ensuring accurate ground truth for each instance. Data augmentation not only increases the quantity of the training augmentation data but also improves the robustness of the model by presenting it with a variety of scenarios and conditions.

Finally, the point cloud was downsampled to achieve an initial count of 11,000 points for each point cloud. This number was dictated by the system's capacity to effectively train the network. To optimize the process, a varying downsampling rate was applied based on the nature of the objects: static objects like walls and floors were downsampled at a rate of 10:1 (every ten points, one was saved), whereas dynamic objects were downsampled at a rate of 2:1. This differentiated approach ensures that more critical dynamic interactions are captured in greater detail, enhancing the training and eventual performance of this models in real-world scenarios.





\subsection{\textbf{Network Architecture}}
In this section, a novel dual-stream network named FUSION is introduced, which is designed to address point cloud semantic segmentation and perform domain adaptation tasks. 
This network uniquely combines the strengths of a Dynamic Graph Convolutional Neural Network (DGCNN) \cite{DGCNN} and a Convolutional Neural Network equipped with residual connections (CNN-residual).

\begin{figure*}[] 
\centering
\includegraphics[width=1.1\textwidth]{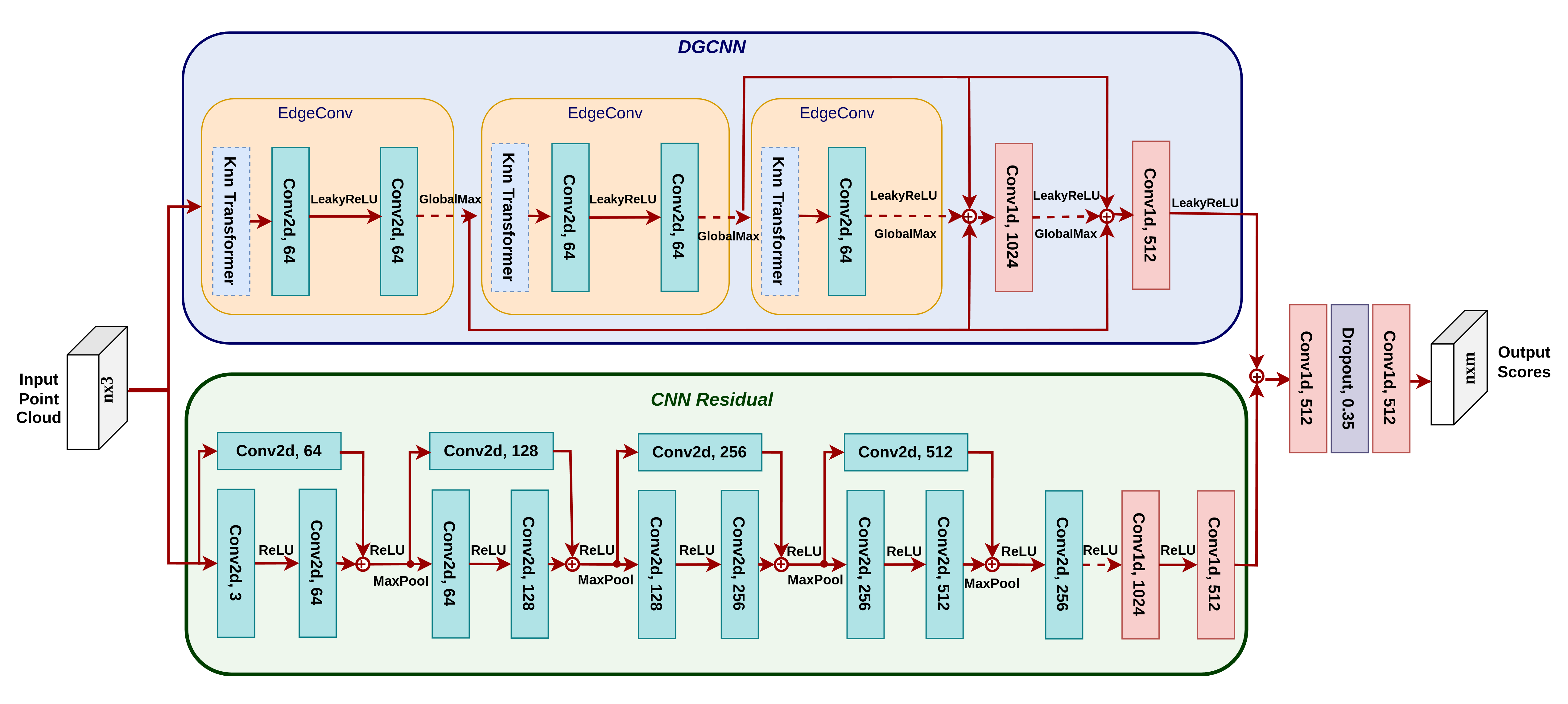}
\caption{The FUSION Network: a dual-stream network architecture combining DGCNN and CNN-residual.}
\label{fig:architecture}
\vspace{-3 mm}
\end{figure*} 

The first stream leverages the power of DGCNN for point cloud data. The edge convolution operation (EdgeConv) from DGCNN, known for its capability to capture local geometric structures by considering relations between a point and its neighbors, is ideally suited for point cloud data. Given a point \( p \) in a point cloud \( P \) and its \( k \)-nearest neighbors determined by knn, EdgeConv considers the edge features formed by the relationship between a point and its neighbors. Mathematically, the edge feature \( h \) for point \( p \) and its neighbor \( q \) is given by

\begin{equation}
h(p, q) = \phi(p - q) \label{eq1}
\end{equation}

where \( \phi \) is a function that encodes the relative spatial position between the points. In this network, the EdgeConv operation is defined as:

\begin{equation}
h_i^{\prime} = \max_{j:(i,j) \in \mathcal{E}} \Theta \cdot (h_i \oplus (h_j - h_i)) \label{eq2}
\end{equation}

where \( h_i \) and \( h_j \) are feature vectors of points \( i \) and \( j \) respectively, \(\mathcal{E}\) denotes the edges in the graph, \(\Theta\) represents the learnable parameters of the network, and \(\oplus\) denotes concatenation. 
The graph structure is adaptively updated at each layer, allowing for the computation of neighborhood information to be dynamic. This dynamic graph-based approach effectively learns geometric and spatial features, enabling a robust understanding of complex point cloud structures.

It also offers invariance to certain transformations like translations and rotations. It allows the model to capture and learn from the intricacies of the point cloud data, such as varying density and irregular formats, that are often overlooked by traditional CNNs.

Complementing the DGCNN, the second stream incorporates a CNN with residual connections. The integration of residual connections aids in learning both low-level and high-level feature hierarchies, resulting in an enriched understanding and processing of the input data across a variety of datasets. Furthermore, the residual connections are added around every two layers which help mitigate the vanishing gradient problem, enabling the network to learn deeper representations effectively.

\begin{equation}
y = F(x) + x
\end{equation}

Here, $F(x)$ represents the residual mapping learned by the network, and $x$ is the input to the residual block. This structure allows the gradient to flow through the network more effectively, thereby addressing the vanishing gradient problem. Specifically, in the CNN residual stream, each residual block consists of two convolutional layers:

\begin{equation}
F(x) = \text{ReLU}(\text{Conv2d}(\text{ReLU}(\text{Conv2d}(x))))
\end{equation}

By combining DGCNN and CNN-residual, the proposed network achieves superior performance in both semantic segmentation of point clouds and the domain adaptation task in the industrial use case. By learning the underlying geometric structures and exploiting residual connections, this network offers a powerful solution for point cloud processing tasks. It enables efficient knowledge transfer between synthetic and real-world datasets, paving the way for the practical deployment of robotics applications in diverse real-world scenarios.

Figure ~\ref{fig:architecture} illustrates the proposed architecture for Fusion. This figure visually summarizes the dual-stream nature of the network and offers an intuitive understanding of its functionality and it demonstrates how the DGCNN and CNN-residual parts work together. 
Each 2D convolutional layer is paired with a batch normalization operation for improved training stability and faster convergence. 
Using ReLU in the residual stream enables efficient training with strong non-linearities and prevents vanishing gradients, while employing Leaky ReLU in the DGCNN stream ensures robust learning from dynamic graph-structured data by preventing neuron saturation and enabling effective handling of varying input distributions.

\section{Experimental Results}\label{sec:Experimental Results}

This section present the results of deploying the network on real-world data, benchmarking its performance against current state-of-the-art algorithms. A fair comparison is ensured by standardizing the testing conditions across all evaluated methods. Ultimately, the network was also tested in real-time scenarios to demonstrate its practical applicability and effectiveness in dynamic environments. This testing aims to validate the robustness and reliability of the proposed approach under both controlled and live conditions.
\subsection{\textbf{Experimental Setup}}

In the experimental setup, 80\% of the data was allocated for training and 20\% for evaluation. Additionally, in the testing phase, 25 samples were utilized for fine-tuning and 81 real-world data samples for final testing.
The system setup included an NVIDIA RTX 8000 GPU paired with an Intel® Xeon(R) Gold 6256 CPU, providing a robust foundation for training this complex models. This configuration facilitated smooth training processes with minimal delays. 

During the training process, the cross-entropy loss was used as a segmentation loss to guide the network to the optimal solution. The Adam (Adaptive Moment Estimation) optimizer was employed due to its efficiency and low memory requirements. The learning rate is set initially to 0.001 and is decreased during the training process based on a learning rate schedule (Cosine decay).

\subsection{\textbf{FUSION Performance}}

In the experimental results, a phenomenon was observed that aligns with a broader challenge in the field, known as the 'sim-to-real gap'.
This gap becomes particularly apparent when the performance of machine learning models trained on simulated datasets is compared with their effectiveness in real-world settings. The gap is due to differences in factors like environmental complexity, data disparity, sensor accuracy, lighting conditions, object variability, etc. Such discrepancies highlight the challenge of applying simulated models to real-world situations.

Firstly, the network was trained exclusively using simulated data, achieving an impressive accuracy of 98.9\%.  On deploying this model directly in real-world, this approach resulted in an initial segmentation accuracy of 45 percent. This significant disparity serves as a clear indication of the 'sim-to-real gap'.
To mitigate this divergence, domain adaptation techniques like data augmentation and fine-tuning were employed. Data augmentation introduced diversity into our simulated dataset, making it more representative of real-world complexities. 

In the fine-tuning process, the model improved by learning from a small set of real-world data (25 samples only). It achieved this while retaining knowledge from its original training on simulated data.
What makes this process remarkable is that it resulted in substantial improvements in the model's accuracy by 52.7 percent.

For the evaluation on real-world data, this classification scheme was maintained, ensuring the dataset comprised the same seven classes: Robot, Human, Wall, Floor, Assembly Line, Table, and AGV (Automated Guided Vehicle). An important aspect of this fine-tuning strategy was the decision to freeze and fix the parameters of the preceding layer, except for the last two fully connected layers. This ensured that these parameters remained unchanged during the retraining and fine-tuning phases.

The advantage of freezing the parameters in the previous layer lies in its ability to preserve the specialized knowledge acquired during the initial training while enabling the model to acquire fresh insights for the specific target domain.

When the parameters of the previous layer are frozen, they are protected from further training or changes. This ensures that the model doesn't forget what it has already learned. At the same time, it gives the model the opportunity to focus on acquiring fresh insights that are specifically tailored to the current task. This maximizes the model's use of past experiences while allowing it to be adaptable to new challenges.

The results of these efforts were indeed remarkable. The model achieved 97.92 percent accuracy when assessed using real-world data containing the seven classes and 97.67 percent accuracy when considering the complete set of eight classes. This underlines the efficacy of the fine-tuning process and its ability to adapt the model to real-world scenarios, yielding highly accurate outcomes.

This experience highlights the importance of employing sim2real gap mitigation techniques to improve adaptability of machine learning models for real-world challenges. It demonstrates a true potential to bridge the gap between simulation and reality, allowing models to excel in complex real-world.

\subsection{\textbf{Performance Evaluation}}

In this study, the performance of the FUSION algorithm was compared with several state-of-the-art algorithms to ascertain its effectiveness across various metrics. In evaluating the performance, it was not only examined key metrics such as Overall Accuracy, Per-class Accuracy, and Mean Intersection over Union (IoU) but also considered the computational time, which is a crucial aspect of practical applicability. The Overall Accuracy (OA) serves as a primary indicator, measuring the proportion of correctly predicted data points out of the total dataset.

\begin{table*}[]
\centering
\caption{Class IOUs for Different Algorithms}
\begin{tabular}{lcccccccc}
\toprule
\multirow{2}{*}{Algorithm} & \multicolumn{7}{c}{Class IOUs} \\
\cmidrule{2-8}
 & Robot & Human & Wall & Floor & Assembly line & Table & AGV \\
\midrule
RandLaNet & 0.930 & 0.930 & 0.953 & 0.937 & 0.957 & 0.880 & \textbf{0.992}\\
PointNet & 0.858 & 0.692 & 0.904 & 0.899 & 0.914 & 0.870 & 0.545 \\
DGCNN & 0.912 & 0.699 & 0.957 & 0.908 & 0.959 & 0.853 & 0.571\\
CNN-Residual & 0.909 & 0.829 & 0.966 & 0.924 & 0.960 & 0.841 & 0.772\\
FUSION(Ours) & \textbf{0.955} & \textbf{0.943} & \textbf{0.974} & \textbf{0.947} & \textbf{0.978} & \textbf{0.923} & 0.922 \\
\bottomrule
\end{tabular}
\label{tab:class_ious}
\vspace{-1 mm}
\end{table*}

\begin{table}[]
\centering
\caption{Algorithm Comparison}
\begin{tabular}{lcccccc}
\toprule
\multirow{2}{*}{Algorithm} & \multicolumn{3}{c}{Metrics} & \multirow{2}{*}{Time (s)} \\
\cmidrule{2-4}
 & Overall Acc. & Per-class Acc. & mIOU & \\
\midrule
RandLaNet & 96.34 & - & 0.944 & 0.281 \\
PointNet & 91.17 & 88.69 & 0.856 & 0.163 \\
DGCNN & 93.15 & 89.94 & 0.881 & 0.090 \\
CNN-Residual & 94.45 & 87.84 & 0.904 & 0.035 \\
FUSION(Ours) & \textbf{97.76} & \textbf{96.82} & \textbf{0.954} & 0.115 \\
\bottomrule
\end{tabular}
\label{tab:algorithm_comparison}
\vspace{-2mm}
\end{table}

Intersection over Union (IoU) measures the overlap between two sets, often used in context of object detection or segmentation:
\begin{equation}
IoU = \frac{Area\_of\_Overlap}{Area\_of\_Union} = \frac{TP}{TP + FP + FN} \label{eq2}
\end{equation}

Where:

     TP = True Positives(correctly predicted points).
     
     FP = False Positives(incorrectly predicted points).
     
     FN = False Negatives(missed points)

\vspace{\baselineskip}

Mean Intersection over Union (mIoU) is the average IoU across multiple classes or instances:

\begin{equation}
mIoU = \frac{1}{N} \sum_{i=1}^{N} IoU_i
\label{eq:meanIoU}
\end{equation}

Where:
\begin{align*}
& N = \text{Number of classes or instances} \\
& IoU_i = \text{IoU for class or instance } i
\end{align*}

\begin{figure*}[] 
\centering
\hspace{8mm}
\includegraphics[width=1.1\textwidth]{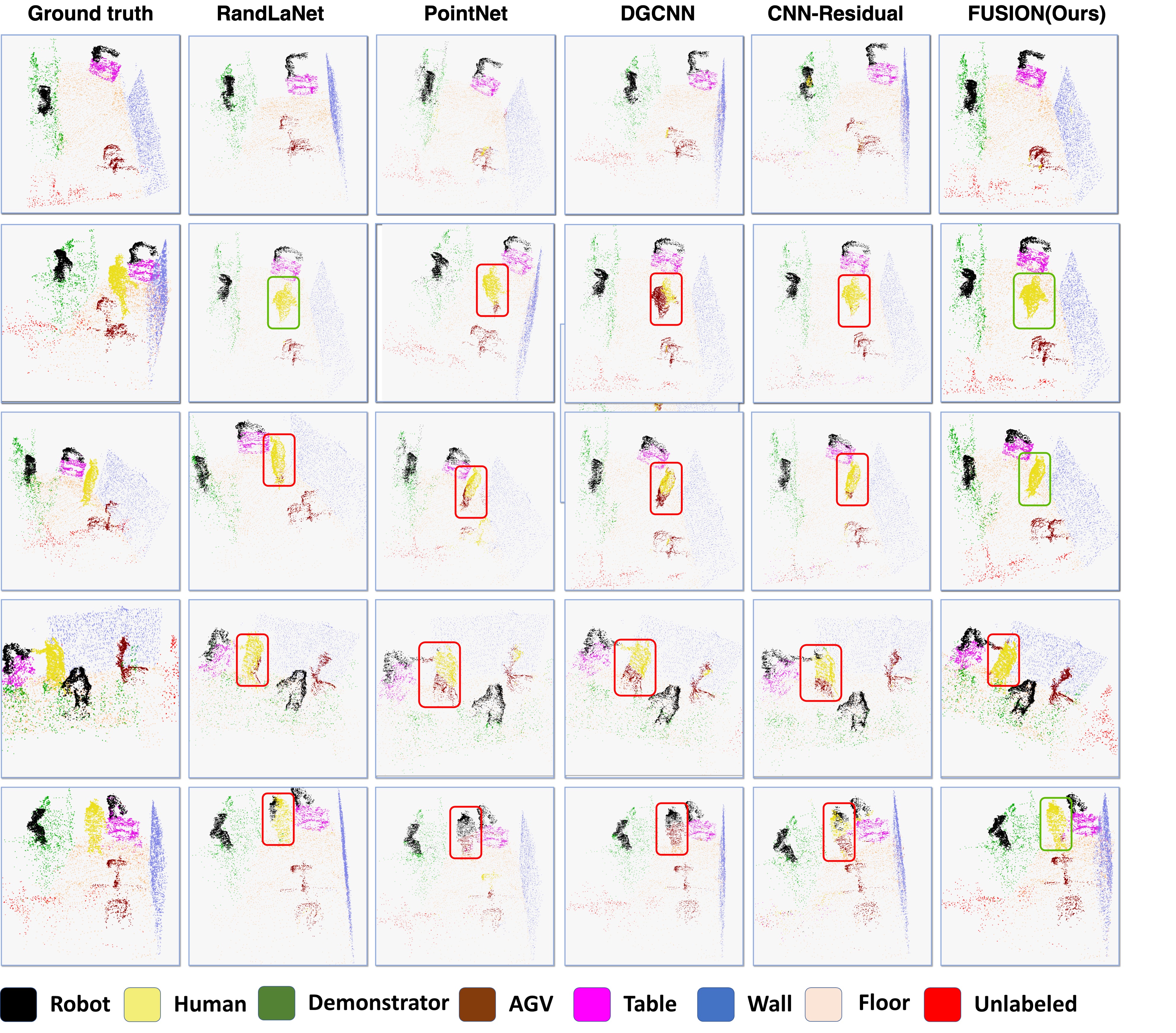}
\caption{Segmentation results for each network, the comparison human segmentation; the green rectangle signifies a successful and complete segmentation, while a red rectangle indicates subpar or inaccurate segmentation.}
\label{fig:seg1}

\end{figure*}

The results, presented in Table \ref{tab:class_ious} and Table \ref{tab:algorithm_comparison}, provide a comparative analysis of these metrics for different algorithms. 
In Table \ref{tab:algorithm_comparison}, a comprehensive comparison of various algorithms is conducted to assess their performance in Sim2Real segmentation tasks. 
As shown in Figure \ref{fig:seg1}, it is evident that the proposed network outperforms all the other approaches across all categories, especially the human body, apart from the AGV segmentation, where the RandLaNet algorithm is marginally better \major{(The 'Unlabeled' class represents regions within point clouds that lack clear structural information or relevance to the segmentation classes)}. Although RandLaNet slightly outperforms FUSION on AGV segmentation, this can be attributed to its efficient multi-scale sampling and spatial consistency, which better capture rigid structures like AGVs. In contrast, FUSION is optimized for complex, detail-rich objects such as humans and robots.

This human body segmentation is crucial, especially in industrial scenarios where precise human body segmentation can be vital for safety as is the case for human-robot collaboration in flexible automation processes.
These results show that proposed model has immense potential in the world of Sim2Real segmentation, especially in scenarios demanding precise human body detection.

Moreover, This model demonstrates a significant improvement in processing speed, operating at just 40\% of the prediction time required by the top competing algorithm (RandLaNet).
This difference is important in HRC contexts where rapid processing and decision-making are essential for smooth and safe human-robot interactions. This FUSION approach not only maintains competitive accuracy but also excels in operational efficiency, making it more suitable for real-time applications where quick response times are critical.

\subsection{\textbf{Real-Time Results}}

\begin{figure*}[] 
\centering
\includegraphics[width=0.94\textwidth]{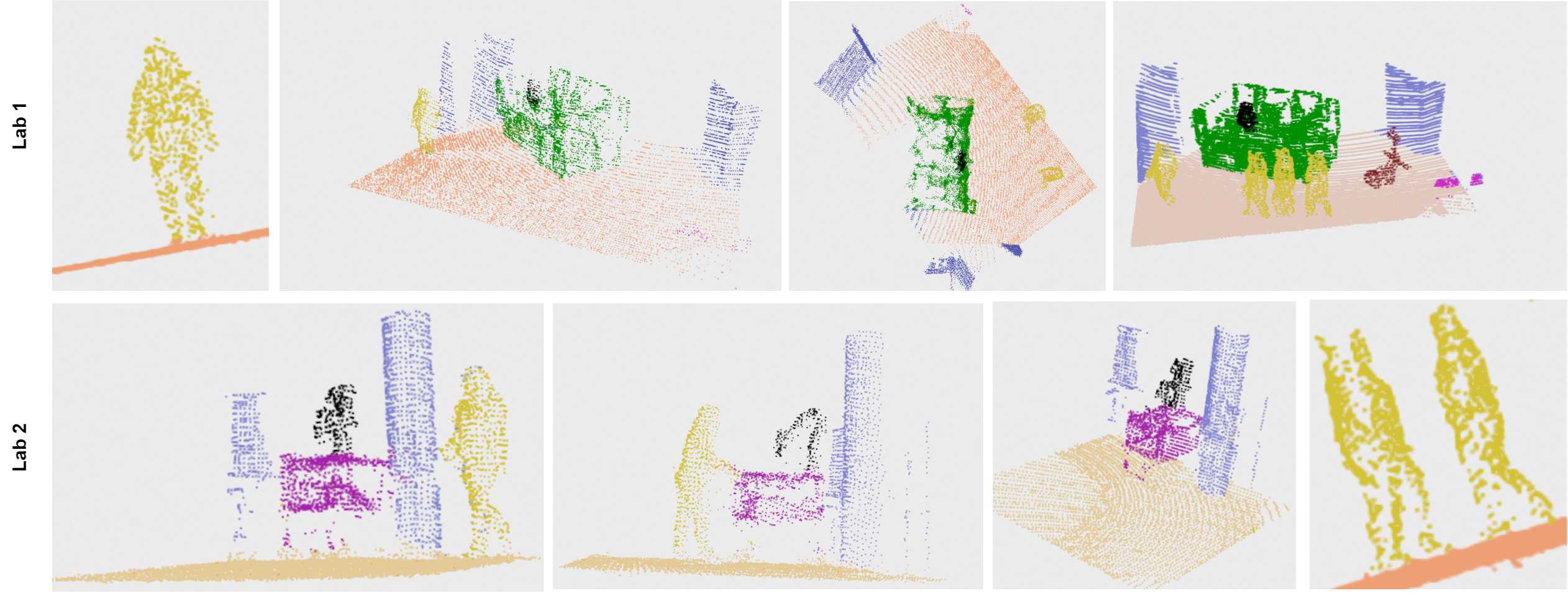}
\caption{Real-Time segmentation results in the HRC industrial application, two Lab setups}
\label{fig:Real}
\hspace{-2mm}
\end{figure*}

In this study, the performance of FUSION network in real-time scenarios is evaluated using two different laboratory setups, each equipped with two LiDAR sensors. To enhance the comprehensiveness of point clouds, the data from both LiDAR sensors are merged, resulting in more complete point cloud representations. 
The obtained results were promising, demonstrating the network's efficacy in processing real-world data. It was observed that the prediction time, coupled with visualization, averaged about 0.19 seconds(
please refer to this video \footnote{\href{https://youtu.be/nPzw3DoRVYA}{https://youtu.be/nPzw3DoRVYA}} for segmentation visualization before optimization). 

Visualization is recognized as a time-intensive component that is not always necessary in real-world scenarios. To optimize this process, a strategy was implemented to exclude static objects such as floors and walls from ongoing visualizations after their initial segmentation. 

Furthermore, downsampling was implemented with a voxel size of 0.05, successfully reducing the processing time to 0.098 seconds, with segmentation and preprocessing taking 0.073 seconds
(please refer to this video \footnote{\href{https://youtu.be/0OLgQdzqwvQ}{https://youtu.be/0OLgQdzqwvQ}} for segmentation visualization after optimization).
This optimization not only improved efficiency but also highlighted the adaptability of the proposed approach to dynamic real-world environments.

Figure  \ref{fig:Real} provides a comprehensive visual representation of the segmentation results obtained from the FUSION network across both laboratory setups. These results are crucial for evaluating the network's performance and its ability to accurately distinguish various objects and features within the environment. In Lab 1, the segmentation results reveal the precise identification of objects such as Human, Robot, AGV and Assembly line, showcasing the network's effectiveness in distinguishing between different classes. Notably, the segmentation results in both labs demonstrate consistency and reliability, highlighting the robustness of FUSION network across diverse real-world settings. These visualizations provide valuable insights into the network's capabilities and serve as a crucial validation of its effectiveness in real-time applications.

\section{Conclusion and Outlook}\label{sec:conclusion}

\major{This study addresses the challenge of Sim2Real domain adaptation in the context of semantic segmentation of 3D point cloud data within industrial environments, with a particular emphasis on real-world performance. Industrial settings are inherently complex, characterized by diverse objects, dynamic configurations, and unpredictable movements, which require models capable of robust real-world performance.}

\major{The research identifies a significant gap in bridging the domain shift between simulated and real-world data which is a major obstacle in deploying machine learning models effectively in safety-critical industrial settings like HRC scenarios. }

\major{To tackle this gap, this research proposed the FUSION network, a dual-stream architecture combining the strengths of Dynamic Graph Convolutional Neural Networks (DGCNN) and Convolutional Neural Networks with residual connections (CNN-residual). This approach utilized simulated data for initial training, followed by adaptation for real-world deployment.
Comprehensive evaluations on both simulated and real-world industrial datasets demonstrated the network's capabilities. }

\major{The decision to use a dual-stream design was based on the unique strengths of the two selected networks. The DGCNN effectively captured local geometric details, while the CNN-residual can handle both basic features like edges and textures, as well as high-level feature hierarchies, such as object representations, and it helps with the vanishing gradient problem. This combination ensures superior performance during the challenging transition from simulated to real conditions.}

\major{To sum up, this paper contributes towards bridging the gap between simulated and real-world industrial setups using the "FUSION" dual-stream network design. The results in practice, despite the challenges in this domain, have shown that the proposed approach has great potential and outperforms previous works. As the field of industrial domain adaptation keeps evolving, this architecture provides a strong and adaptable foundation for future improvements.}

\section{Limitations and Future Work} 

While this study shows promising results, it is important to recognize certain limitations that we aim to address in the future. We used the Nvidia IsaacSIM platform, which is valuable for simulating industrial scenarios but challenging in replicating real-world dynamic aspects accurately.  This limitation becomes especially apparent in scenarios involving interactions between robots, humans, and dynamic objects.
Notably, the software could not render the point cloud for dynamic objects. Consequently, we utilized a humanoid object as a surrogate for a real human and made manual adjustments to the robot's path planning in each simulation.

\major{Future work will focus on enhancing simulation capabilities or exploring alternative platforms to replicate real-world dynamics better, enabling the integration of more diverse and dynamic elements into our dataset. Additionally, safety-critical scenarios involving human-robot interactions where the safety of human workers collaborating with robots is paramount will be prioritized, particularly in modeling and mitigating potentially dangerous situations.}

\major{To improve efficiency, we aim to parallelize the dual-stream model’s operations, significantly reducing segmentation time and improving responsiveness. Exploring deeper integration of the two network streams, attention mechanisms for refined feature extraction, and expanding the dataset to cover a broader range of industrial settings will further enhance the robustness and applicability of our approach.}

\bibliography{biblography}

\section*{Declarations}

\subsection{\textbf{Funding}}

This work was partially supported by DIZH (Digitalization Initiative of the Zurich Higher Education Institutions) funding.

\subsection{\textbf{Conflict of interest}}

The authors have no relevant financial or non-financial interests to disclose.

\subsection{\textbf{Consent to publish}}
The authors affirm that human research participants provided informed consent for the publication of the images in Figure 2b.

\subsection{Data Availability}
A portion of the dataset used in this study is publicly available and can be accessed at \href{https://github.com/Fatemeh-MA/COVERED}{https://github.com/Fatemeh-MA/COVERED}. The simulation dataset and the code will be made publicly available upon acceptance of this paper.

\end{document}